\documentclass[letterpaper, 10 pt, journal, twoside]{ieeetran}

\usepackage{myStyle}

\begin{document}

\title{Sampling-based Model Predictive Control Leveraging Parallelizable Physics Simulations}

\author{Corrado Pezzato*, Chadi Salmi*, Elia Trevisan*, Max Spahn, Javier Alonso-Mora, Carlos Hern\'andez Corbato

\thanks{$^*$ Equal contribution, alphabetical order.}
\thanks{This research was supported by Ahold Delhaize and the ``Sustainable Transportation and Logistics over Water: Electrification, Automation and Optimization (TRiLOGy)'' project of the Netherlands Organization for Scientific Research (NWO), domain Science (ENW).}
\thanks{The authors are with the Cognitive Robotics Department,
        TU Delft, 2628 CD Delft, The Netherlands
        {\tt\small \{c.pezzato, c.salmi, e.trevisan, m.spahn, j.alonsomora, c.h.corbato \}@tudelft.nl}}
}

\markboth{IEEE ROBOTICS AND AUTOMATION LETTERS. ACCEPTED VERSION.}
{Pezzato \MakeLowercase{\textit{et al.}}: Sampling-based Model Predictive Control Leveraging Parallelizable Physics Simulations} 

\maketitle

\IEEEpeerreviewmaketitle


\begin{abstract}
We present a sampling-based model predictive control method that uses a generic physics simulator as the dynamical model. In particular, we propose a Model Predictive Path Integral controller (MPPI) that employs the GPU-parallelizable IsaacGym simulator to compute the forward dynamics of the robot and environment. Since the simulator implicitly defines the dynamic model, our method is readily extendable to different objects and robots, allowing one to solve complex navigation and contact-rich tasks. We demonstrate the effectiveness of this method in several simulated and real-world settings, including mobile navigation with collision avoidance, non-prehensile manipulation, and whole-body control for high-dimensional configuration spaces. This is a powerful and accessible open-source tool to solve many contact-rich motion planning tasks. Code and videos: \url{https://autonomousrobots.nl/paper_websites/isaac-mppi}
\end{abstract}


\section{Introduction}
\label{sec:intro}
\IEEEPARstart{A}{s} robots become increasingly integrated into our daily lives, their ability to navigate and interact with the environment is becoming more important than ever. From collision avoidance to moving obstacles out of the way to pick up some objects, robots must be able to plan their motions while accounting for contact with their surroundings. At the same time, robotic platforms require many Degrees Of Freedom (DOF) to achieve agile and dexterous movements. All this poses many challenging problems to motion planners, such as collision-free navigation in complex and dynamic environments, high DOF mobile manipulation, contact-rich tasks such as picking and pushing, and in-hand manipulation. 
Solutions to these challenges exist but are often specialized and not easily transferable to different scenarios. Learning-based approaches, for example, can leverage physics simulators to train policies for complex tasks but require extensive training and resources. For instance, \cite{andrychowicz_learning_2020} took years to develop, utilizing 6144 CPU cores and 50 hours of training to learn a policy for in-hand cube manipulation.
\begin{figure}[t!]
    \centering
    \includegraphics[width=1.0\linewidth]{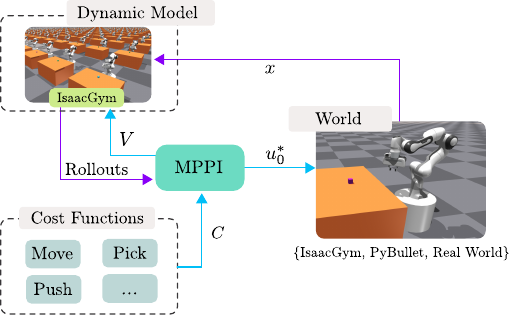}
    \caption{Scheme of the proposed method using IsaacGym as the dynamic model for MPPI. At each time step, IsaacGym is reset to the current world's state $x$, and random input sequences ${V}$ are applied for the horizon $T$, to every environment. MPPI uses the resulting rolled-out trajectories to approximate the optimal control $u_0^*$ given a cost function $C$.}
    \label{fig:general_idea}
    \vspace{-2mm}
\end{figure}
On the other hand, model-based approaches like Model Predictive Control (MPC) can solve challenging tasks \cite{schwenzer_review_2021}. \textcolor{black}{However, MPC often relies on constrained optimization, requiring constraint simplifications, precise modeling, and ad-hoc solutions to handle discontinuous dynamics in contact-rich tasks \cite{hogan_reactive_2020, moura_non-prehensile_2022}. While utilizing motion memory for warm-starting optimization can enhance performance \cite{mansard_using_2018, lembono_memory_2020}, the above limitations still persist.}
Recently, Model Predictive Path Integral (MPPI) control~\cite{williams_model_2017} and its information-theoretic counterpart~\cite{williams_information-theoretic_2018} addressed optimal control problems via importance sampling, mitigating challenges tied to constrained optimization algorithms dealing with non-convex constraints and discontinuous dynamics. However, substantial modeling remains necessary.

In this paper, we propose a training-free model-based framework for real-time control of complex systems, where one designs only a cost function, not the problem's dynamics and contact models. We introduce the idea of using a general GPU-parallelizable physics simulator, IsaacGym~\cite{makoviychuk_isaac_2021}, as the dynamic model for MPPI. This creates a robust framework that generalizes to various tasks. An overview is given in \Cref{fig:general_idea}.

\subsection{Related work}
\label{sec:related_work}

This section provides an overview of selected works focusing on motion planning and contact-rich tasks in robotics. Motion planning pipelines are categorized as global and local motion planning \cite{siegwart_introduction_2011}. Local motion planning encompasses approaches like operational space control, geometric methods such as Riemannian Motion Policies \cite{li_multi-objective_2022} and Optimization Fabrics \cite{ratliff_generalized_2021,spahn_dynamic_2023}, and receding-horizon optimization formulations like Model Predictive Control (MPC) \cite{buizza_avanzini_constrained_2018} that may incorporate learned components \cite{hewing_learning-based_2020}. Most MPC algorithms rely on constrained optimization and assume smooth dynamics. However, contact-rich tasks pose challenges due to their non-smooth and hybrid nature, involving sticking and sliding frictions or entering contacts, requiring extensive modeling and ad-hoc solutions for pushing tasks~\cite{hogan_reactive_2020, moura_non-prehensile_2022}.

In contrast, Model Predictive Path Integral (MPPI) control \cite{williams_model_2017, williams_information-theoretic_2018} is a sampling-based MPC approach that approximates optimal control via parallel sampling of input sequences. MPPI is gradient-free and well-suited for systems with non-linear, non-convex, discontinuous dynamics and cost functions. It has successfully controlled high-degree-of-freedom manipulators in real-time \cite{bhardwaj_storm_2022}, incorporating self-collision avoidance using trained neural networks and collision-checking functions \cite{danielczuk_object_2021}. However, these approaches have limited interaction with the environment. In \cite{abraham_model-based_2020}, the authors propose ensemble MPPI, a variation that handles complex tasks and adapts to parameter uncertainty. Still, the task modeling remains unclear, and no open-source implementation is available.

To alleviate the problem of explicit modeling, some works have addressed the use of physics simulators for sampling-based MPC. In \cite{carius_constrained_2022}, the authors use the RaiSim simulator to sample waypoints for foot placement of a quadruped. Moreover, Howell et al. \cite{howell_predictive_2022} proposed a sampling-based MPC method that employs MuJoCo~\cite{todorov_mujoco_2012} as a dynamic model for rolling out sampled input sequences. This offloads modeling efforts to the physics engine, simplifying controller design. However, MuJoCo's parallelization capabilities are constrained by the number of CPU threads, limiting real-time performance when many samples are required to solve a task. \textcolor{black}{Moreover, results are presented only in simulation.}

A high number of samples is particularly crucial in tasks such as non-prehensile manipulation with robot manipulators. Traditional approaches often involve sampling end-effector trajectories on a plane, relying on additional controllers for robot actuation and learned models for predictions \cite{arruda_uncertainty_2017, cong_self-adapting_2020}. For instance, Arruda et al. \cite{arruda_uncertainty_2017} use a forward-learned model trained on 326 real robot pushes. This model is employed by an MPPI controller to plan push manipulations as end-effector trajectories. Cong et al. \cite{cong_self-adapting_2020} train a \textcolor{black}{Long Short-Term Memory-based} model to capture push dynamics using a dataset of 300 randomized objects. End-effector trajectories are sampled within a rectangular 2D workspace. Both methods require a separate controller to convert cartesian motions into joint commands, and both perform push manipulation through a sequence of pushes, resulting in discontinuous motion. These methods are not easily transferable to other robots, particularly non-holonomic mobile pushing.

\subsection{Contributions}
This paper presents a novel open-source implementation of Model Predictive Path Integral (MPPI) control with a generic physics simulator as the dynamical model. This enables the method to solve many contact-rich motion planning problems. The two key contributions of this work are: 
\begin{itemize}
    \item The integration of the MPPI controller with the GPU-parallelizable simulator  IsaacGym, distinguishing our approach from prior works in MPPI. Our method facilitates collision checking and contact-rich manipulation tasks leveraging the contact models and rigid body interactions included in the simulator without requiring gradients. Our solution allows smooth real-time control of real-world systems with high degrees of freedom, efficiently computing hundreds of rollouts in parallel. 
    \item A versatile method applicable to various motion planning challenges, including collision avoidance, prehensile and non-prehensile manipulation, and whole-body control with diverse robots. We provide an open-source implementation that can be readily reused and extended to heterogeneous robots and tasks.
\end{itemize}
We perform many contact-rich tasks with several robotic platforms and real-world experiments. We include omnidirectional and differential drive robots and fixed or mobile manipulators and compare against many specialized baselines.

\section{Sampling-based MPC via parallelizable physics simulations}


In this section we describe the integration of MPPI with IsaacGym, which enables real-time control of complex contact-rich robotic systems with minimal modeling.


\subsection{Background theory on MPPI}

In this section, we give an overview of the background theory of MPPI. For more theoretical insights, please refer to the original publications~\cite{williams_model_2017, williams_information-theoretic_2018}. MPPI is a method to solve stochastic optimal control problems for discrete-time dynamical systems such as
\begin{equation}
    {x}_{t+1} = {f}({x}_t, {v}_t),\ \ \ {v}_t \sim \mathcal{N}(u_t, \Sigma),
\end{equation}
where the nonlinear state-transition function ${f}$ describes how the state $x$ evolves over time $t$ with a control input ${v}_t$. MPPI samples $K$ noisy input sequences ${V}_k$. These sequences are then applied to the system to simulate $K$ state trajectories $Q_k$, $k \in [1,K]$, over a time horizon $T$:
\begin{equation}
    Q_k = [{x}_0, {f}({x}_0, {v}_0), \dots, {f}({x}_{T-1}, {v}_{T-1})].
\end{equation}

Given the state trajectories $Q_k$ and a designed cost function $C$ to be minimized, the total state-cost $S_k$ of an input sequence $V_k$ is computed by functional composition $S_k = C(Q_k)$. Then, each rollout is weighted by importance sampling weights $w_k$, computed via an inverse exponential of $S_k$ with tuning parameter $\beta$,  normalized by $\eta$. The minimum sampled cost $\rho = \min_k S_k$ is subtracted for numerical stability, leading to:
\begin{equation}
    \label{eq:weights}
    w_k = \frac{1}{\eta}\exp\left(-\frac{1}{\beta}(S_k - \rho)\right), \ \ \ \sum_{k=1}^K w_k=1
\end{equation}
The parameter $\beta$ is also known as \textit{inverse temperature}. The weights are then used to compute the approximate optimal control input sequence $U^*$:
\begin{equation}
    \label{eq:approx_U}
    U^* = \sum_{k=1}^K w_k {V}_k
\end{equation}
\Cref{eq:weights} and \Cref{eq:approx_U} demonstrate the approach taken to approximate the optimal control. \Cref{eq:approx_U} represents a weighted average of sampled control inputs, while \Cref{eq:weights} assigns exponentially higher weights to less costly inputs.
The first input $u_0^*$ of the sequence $U^*$ is applied to the system. Then the process is repeated.

\subsection{Proposed algorithm}
We now describe how we use MPPI with IsaacGym, summarized in \Cref{alg:mppi_isaacgym}. We initialize an input sequence $U_{init}$ as a vector of zeroes with a length of $T$, where $T$ is the time horizon in steps. 
We then sample $K$ sequences of additive input noise $\mathcal{E}_k$ for exploring the input space around $U_{init}$. The key concept is that, instead of explicitly defining a nonlinear transition function ${f}$, we use IsaacGym to compute the next state ${x}_{t+1}$ given ${x}_t$ and control input ${v_t}$.  This is done by reading the current state of the environment, resetting the state of the simulator to the observed values, and then applying the noisy control input sequence to simulate the state trajectories in IsaacGym. Note that these $K$ state trajectories can be computed independently of each other. We use this property to forward and simulate all the rollouts in parallel, leveraging the parallelization capabilities of IsaacGym.
Instead of sampling from a Gaussian distribution, we follow the strategy of a recent paper~\cite{bhardwaj_storm_2022} that proposes to sample \textit{Halton Splines} instead for better exploration and smoother trajectories. 
\textcolor{black}{Similar to~\cite{bhardwaj_storm_2022}, we fit B-Splines to inputs sampled from a Halton sequence using standard Python modules}
and then we evaluate the spline at regular intervals to \textcolor{black}{retrieve} $\mathcal{E}_k$. Unlike~\cite{bhardwaj_storm_2022}, we do not update the variance of the sampling distribution. Instead, we keep it as a tuning parameter, constant during execution. Updating the variance as in~\cite{bhardwaj_storm_2022} can lead to better convergence to a goal, but it also leads to stagnation of the control over time, which is harmful in the contact-rich tasks considered in this paper.
Once the task begins, we reset our $K$ simulation environments on IsaacGym to the current observed \textit{world} state $x$.
In parallel, we can now roll out the sampled input sequences ${V}_k$ into state trajectories $Q_k$ using $K$ simulation environments on IsaacGym and compute their corresponding cost $S_k$ using the designed \textit{cost function} $C$. The cost is discounted over the planning horizon $T$ by a factor $\gamma$~\cite{bhardwaj_storm_2022}:

\begin{equation}
    \label{eq:discounted_cost}
    S_k = \sum_{t=0}^{T-1}\gamma^t C(x_{t, k}, v_{t, k})
\end{equation}

Next, we can compute the \textit{importance sampling} weights $w_k$ as in \Cref{eq:weights}. The normalization factor $\eta$ is a useful metric to monitor, as it indicates the number of samples assigned significant weights. \textcolor{black}{We use this to tune $\beta$ for the next iteration such that $\eta$ is maintained within an upper and lower bound:}
\begin{equation}
    \label{eq:beta_update}
    \beta_{t+1} = 
    \begin{cases}
      0.9\beta_t    & \text{if}\  \eta > \eta_{max} \\
      1.2\beta_t      & \text{if}\  \eta < \eta_{min}\\
      \beta_t      & \text{otherwise}
    \end{cases}
\end{equation}
Empirically, we observed in all performed tasks that setting $5<\eta<10$ is a good balance for smooth behavior. Finally, an approximation of the optimal control sequence $U^*$ can now be computed via a weighted average of the sampled inputs \Cref{eq:approx_U}.
$U_{init}$ is now updated with $U^*$, time-shifted backward of one timestep so that it can be used as a warm-start for the next iteration, \textcolor{black}{$U_{init} = [u^*_1, ..., u^*_{T-1}, u^*_{T-1}] \in \mathbb{R}^T$}. The second last input in the shifted sequence is propagated to the last input as well.
From the sequence $U^*$, only the first input $u_0^*$ is applied to the system, and the next iteration starts.

\begin{algorithm}[th]

\caption{Proposed Approach} \label{alg:mppi_isaacgym}

\begin{algorithmic}[1]
\Initialize{$U_{init}=[0,\dots,0]$ \Comment{$U_{init} \in \mathbb{R}^T$} \\
            $\mathcal{E}_k \gets \textit{sampleHaltonSplines}()$ \Comment{$k=1...K$}}
\While{$\textit{taskNotDone}$}
\State{$x \gets \textit{observeEnvironment}()$}
\State{$\textit{resetSimulations}(x)$}
\For{$k=1 \dots K$} \Comment{in parallel}
\State{${V}_k = U_{init} + \mathcal{E}_k$}
\State{$[Q_k, S_k] \gets \textit{computeRolloutCost}({V}_k, \gamma)$}
\State{$w_k \gets \textit{importanceSampling}$}   \hfill $\triangleright$ \cref{eq:weights}
\EndFor
\State{$\beta \gets \textit{updateBeta}(\beta, \eta)$}    \hfill $\triangleright$ \cref{eq:beta_update}
\State{$U^* = \sum_{k=1}^K w_k {V}_k$}
\State{$U_{init} \gets \textit{timeShift}(U^*)$}
\State{$\textit{applyInput}(u_0^*)$}
\EndWhile
\end{algorithmic}
\end{algorithm}

\subsection{Exploiting the physics simulator features}
IsaacGym provides useful information and general models that are particularly useful for robot control in contact-rich tasks. Besides being useful to simulate the physical interaction of rigid bodies, we leverage IsaacGym for \textit{collision checking} and \textit{tackling model uncertainty} with domain randomization. 
\subsubsection{Collision checking}
Collision checking in robotics can be challenging for a number of reasons, one of them being computational complexity. This is particularly true if the task requires continuous collision checking as the robot moves in dense environments with complex object shapes.  
To overcome this problem, approximations are often introduced with the convexification of the space. However, this requires several heuristics and can hinder robot motions in complex scenes. 

Instead, we propose to tackle the problem of collision checking by using the already available \textit{contact forces tensor} from IsaacGym, which is available for each simulation step. To avoid collisions, we then define a cost function proportional to the contact forces for the MPPI:
\begin{equation}
    C_{coll} = \omega_c \sum F_{obst},
\end{equation}
where $F_{obst}$ are the contact forces exerted on the different obstacles. This allows us to perform continuous collision checking at each time step over the horizon $T$, with arbitrary complex shapes.
By heavily penalizing contacts with obstacles, the robot will avoid collisions. On the other hand, by relaxing the weight $\omega_c$ one can allow for certain contacts required for the task, such as rolling a ball against a wall (Section \ref{sec:omni_sphere}). 

\subsubsection{Tackling model uncertainty}
IsaacGym is designed to easily support domain randomization. We use this feature to randomize the object properties in each environment in case of contact-rich tasks, such that uncertainty is incorporated in every rollout for the MPPI. Effectively, this allows to account
for uncertainty in environment perception.
Specifically, starting from nominal physics properties, in every rollout objects are spawned with uncertainty on mass and friction nominal values, sampled from a uniform distribution.
Additionally, the object size is also randomized with additive Gaussian noise, see \Cref{sec:experiments} for experiment-specific details.
Therefore, every simulation is different from the others, and all simulations are different from the \textit{world} such that we can account for model mismatch. In a sense, we perform a sort of domain randomization in real-time to address the challenge of model uncertainty and imperfect perception.

\section{Experiments}
\label{sec:experiments}
We perform several experiments in three different categories: 1) \textit{motion planning and collision avoidance}, 2) \textit{whole-body control} of high DOF systems in contact-rich settings, and 3) \textit{non-prehensile manipulation}. Experiments and simulations are conducted on an Alienware Laptop with Nvidia 3070 Ti graphics card. The software implementation consists of our \href{https://autonomousrobots.nl/paper_websites/isaac-mppi}{open-source Python package} that can easily be installed, tested, and extended to new robots and tasks. In real-world tests, we used a Robot Operating System (ROS) wrapper to connect the robot to the planner and a motion capture system to determine the pose of manipulated objects.
Our implementation allows for position, velocity, and torque control. In this paper, all robots are velocity-controlled except for the mobile manipulator in~\Cref{sec:whole_body}, which is torque-controlled.

\subsection{Motion planning and collision avoidance}

\begin{figure}[b]
  \centering
  \begin{subfigure}{0.5\linewidth}
    \centering
    \includegraphics[height=2.8cm]{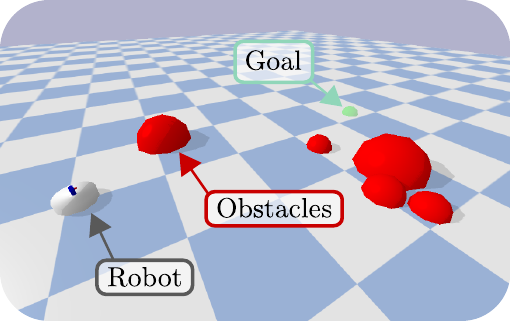}
  \end{subfigure}%
  \begin{subfigure}{0.5\linewidth}
    \centering
    \includegraphics[height=2.8cm]{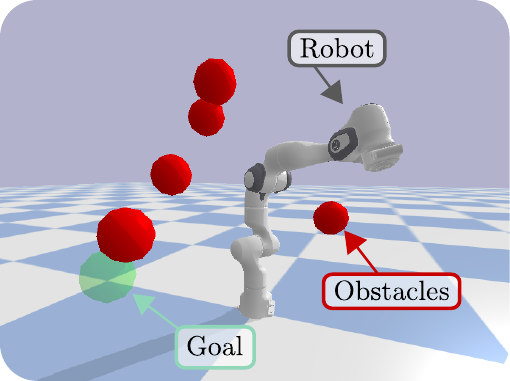}
  \end{subfigure}
  \caption{Examples for pure motion planning benchmark setups. Left: point robot with 3 DOF. Right: manipulator with 7 DOF.
  }%
  \label{fig:benchmark_cases}
\end{figure}

We compare the performance of the proposed method in a pure local motion planning setting, i.e. no interaction with the environment. This aims to showcase the fact that our method is comparable to state-of-the-art techniques when no contact is involved. The main focus is the quantitative analysis of the method compared to two  baselines, 
specifically optimization fabrics as presented in \cite{ratliff_generalized_2021,spahn_dynamic_2023} and a simple MPC formulation solved 
with ForcesPro \cite{zanelli_forces_2020}. We make use of an already available benchmark setup, the \textit{localPlannerBench}\cite{spahn_local_2022}.
We present results for two cases, namely a holonomic robot,
and a robotic arm (Franka Emika Panda). For all experiments, we randomize five obstacles and the goal
positions in $N=100$ runs, see \Cref{fig:benchmark_cases} for some examples. Solutions by the three methods
are assessed using four metrics, e.g. time to reach the goal, path length, solver time, and minimum clearance.

\begin{figure}[t!]
  \centering
  \medskip
  \begin{subfigure}{0.95\linewidth}
    \centering
    \includegraphics[angle=0,width=0.9\linewidth]{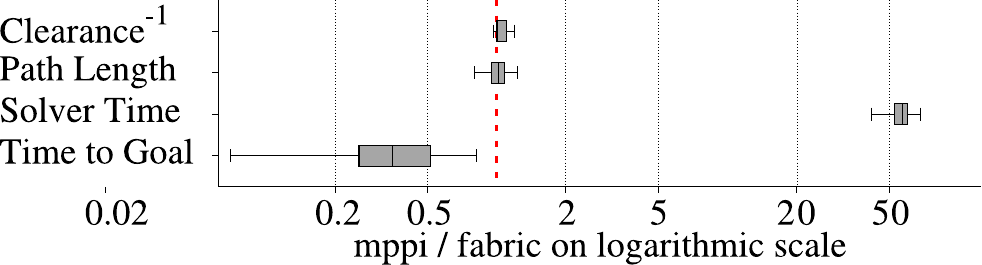}
    \caption{Comparison with Fabrics}%
    \label{subfig:point_results_fabric}
  \end{subfigure}
  \begin{subfigure}{1.0\linewidth}
    \centering
    \includegraphics[angle=0,width=0.9\linewidth]{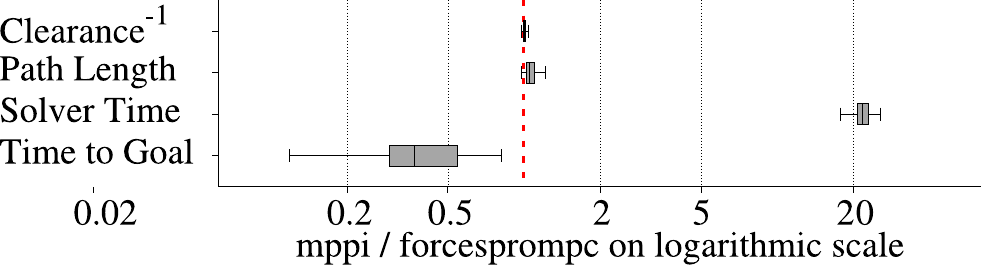}
    \caption{Comparison with ForcesPro MPC}%
    \label{subfig:point_results_mpc}
  \end{subfigure}
  \caption{Results in pure motion planning problems for point robot with 3~DOF.}%
  \label{fig:motion_planning_results_point}
\end{figure}

\begin{figure}[t!]
  \centering
  \begin{subfigure}{0.95\linewidth}
    \centering
    \includegraphics[angle=0,width=0.9\linewidth]{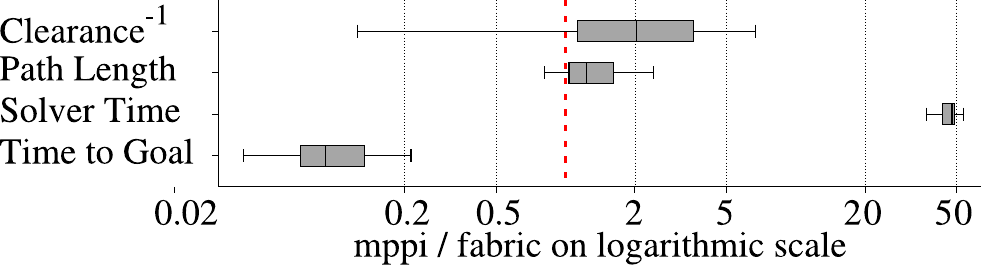}
    \caption{Comparison with Fabrics}%
    \label{subfig:panda_results_fabric}
  \end{subfigure}
  \begin{subfigure}{1.0\linewidth}
    \centering
    \includegraphics[angle=0,width=0.9\linewidth]{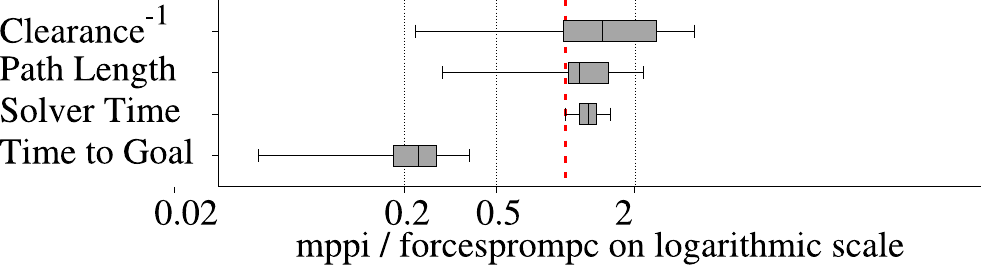}
    \caption{Comparison with ForcesPro MPC}%
    \label{subfig:panda_results_mpc}
  \end{subfigure}
  \caption{Results in pure motion planning problems for a robot manipulator with 7~DOF.}%
  \label{fig:motion_planning_results_panda}
\end{figure}
The compared methods show minimal differences in path length clearance for both examples (\Cref{fig:motion_planning_results_point}). However, our method consistently \textcolor{black}{reaches the goal faster} (\Cref{fig:motion_planning_results_point,fig:motion_planning_results_panda}, and~\Cref{tab:times_planning}). This is attributed to the perfect representation of the robot's collision shapes used in our method, compared to the enclosing spheres in the ForcesPro MPC and optimization fabrics. It should be noted that our approach incurs higher computational times (\Cref{tab:times_planning}) due to the physics simulations performed by IsaacGym. Despite this, our method remains competitive in motion planning applications and offers significant advantages in contact-rich tasks, as demonstrated in the following sections.

\begin{table}[!h]
\caption{Summary of motion planning experiments} \centering 
\begin{tabular}{C{1.8cm} C{0.9cm} C{1.0cm} C{2.0cm} C{1.0cm}} %
\hline\hline 
\textbf{Metric} & \textbf{Robot} & \textbf{Fabric} & \textbf{ForcesPro MPC}& \textbf{MPPI}\\ [0.5ex] 
\hline 
Average & \multicolumn{1}{c}{Point} & \textbf{1.0}ms & 2.5ms & 55ms \\\cline{2-5} 
Solver Time & \multicolumn{1}{c}{Panda} & \textbf{1.4}ms & 51ms & 63ms\\
\hline
Average& \multicolumn{1}{c}{Point}& 7.4s & 6.1s & \textbf{2.7}s \\\cline{2-5}
Time to Goal & \multicolumn{1}{c}{Panda} & 9.6s & 4.2s & \textbf{0.8}s\\
\hline
 \end{tabular}
\label{tab:times_planning}
\end{table}
\vspace{-0.4cm}

\subsection{Prehensile manipulation with whole-body control}
\label{sec:whole_body}
\begin{figure}[htb!]
    \centering
    \medskip
    \includegraphics[width=0.65\linewidth]{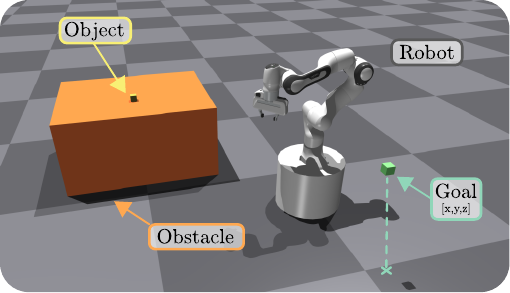}
    \caption{Whole-body motion of a mobile manipulator moving a cube from initial to a desired location.}
    \label{fig:omni_pick}
\end{figure}

Our approach scales well with the complexity of the robot. In \Cref{fig:omni_pick}, the task is to relocate an object from a table to an $[x,y,z]$ location using a mobile manipulator with 12 DOF. 

Although this is arguably a complex task for a robot, which usually requires manual engineering of a sequence of movements, such as navigation to a specific base goal, and pre-post grasps, the solution is rather simple with our method. In fact, we specify the following cost function for the task:  
\begin{equation}
    C_{pick} = C_{dist} + C_{pose} + C_{coll} + C_{vel},
\end{equation}
where we consider the Euclidean distance of the end-effector to the object and the object to the goal: $C_{dist} = \omega_t ||p_{EE} - p_O|| + \omega_{O_p} ||p_G - p_O||$.
We give an incentive to keep the robot in a comfortable pose by penalizing deviations from a desired arm and gripper pose, end-effector orientation, as well as imposing a minimum end-effector height $C_{pose}= C_{Parm}+ C_{Pgrip}+C_{Oee}+ C_{Hee}$. We minimize collisions penalizing the forces on the table $C_{coll}$. Lastly, we penalize high arm and base velocities $C_{vel}= C_{Varm}+ C_{Vbase}$ since, in this experiment, we torque-control the robot.
By sampling all the DOF at once, including the base and the gripper, we achieve a fluid motion from start to end with no added heuristics for pick positions. We performed ten pick-and-deliver tasks, and the time taken was 15.67~$\pm$~7.21s. The high standard deviation is because sometimes the cube falls, but the robot can recover by picking it up again from the floor.

For smooth whole-body motions of high DOF systems like this, many samples are required. Empirically, when the number of samples exceeds 50, a GPU pipeline is computationally cheaper than a CPU and scales better. Using IsaacGym, we can compute all the 750 samples required for mobile manipulation in parallel, computing the next control input online at $25Hz$.
\subsection{Non-prehensile manipulation}
\label{subsec:nonprehensile_sim}
\begin{figure}[htb!]
    \centering
    \includegraphics[width=0.65\linewidth]{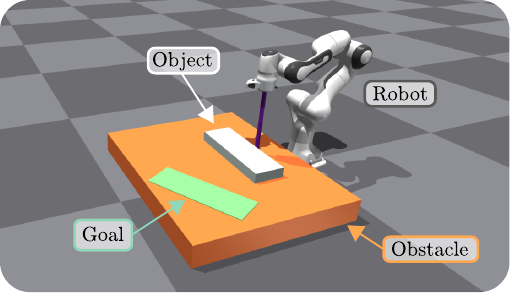}
    \caption{Example of non-prehensile push task with a 7-DOF robot arm using an object from \cite{cong_self-adapting_2020}.}
    \label{fig:pand_push}
\end{figure}

One advantage of using a physics simulator is that one can leverage generic physics rules for contacts, thus eliminating the need for learning or engineering specialized contact models. We demonstrate this in non-prehensile manipulation tasks involving a 7-DOF arm (\Cref{fig:pand_push}) and two different mobile robots (\Cref{fig:non_prehensile_heijn_box,,fig:non_prehensile_heijn_sphere,,fig:non_prehensile_boxer}). In \Cref{subsec:baselines_pushing}, we apply our method to the two \textcolor{black}{pushing} tasks tackled in \cite{arruda_uncertainty_2017, cong_self-adapting_2020}, and we compare with their final results. Additionally, in \Cref{subsec:mobile_pushing}, we demonstrate the ease of transferring our approach to different robots, including differential-drive.

\subsubsection{Comparison with baselines for pushing with a robot arm}
\label{subsec:baselines_pushing}
We consider two baselines for non-prehensile pushing. In the first one, \cite{arruda_uncertainty_2017} tackles the problem of pushing a relatively small object to a target pose with either $0$ (Pose 1) or $90\deg$ (Pose 2). They also consider sequences of push actions starting far from the object. In the second baseline, \cite{cong_self-adapting_2020} considers 5 relatively big objects and assumes the robot's end effector is close to the object during execution. 
Since we do not have access to the same hardware, and the authors of the considered baselines do not provide their models and data, we only compare against their final results. We set up our simulation to match as close as possible the tasks in the baseline using the available information from the papers. Finally, we tune our method for the two tasks separately for a fair comparison with the individual baselines.
The approach in \cite{arruda_uncertainty_2017} utilizes an MPPI in combination with a learned model for predicting pushing effects on an object. The authors sample 2D end-effector trajectories and then rely on inverse kinematic solvers, achieving push manipulation as a sequence of disconnected pushes. In contrast, we use MPPI to sample the control input directly as joint velocities in IsaacGym. By doing so, we achieve smooth continuous pushes where end-effector repositioning emerges naturally, and learning is not required. The cost function to be minimized for the task is:
\begin{equation}
\label{eq:c_push_panda}
    C_{push} = C_{dist} + C_{push\ align} + C_{ee\ align}.
\end{equation}
$C_{dist}$ has the weighted distance \textit{robot-object}, and \textit{object-goal}:
\begin{equation}
\nonumber
    C_{dist} = \omega_t ||p_R - p_O|| + \omega_{O_p} ||p_G - p_O|| + 
    \omega_{O_r} ||\psi_O - \psi_G||,
\end{equation}
where $p_G$ and $\psi_G$ are the goal's position and orientation, while $p_R$ and $p_O$ denote the end-effector tip and block positions, respectively.
The cost function $C_{push\ align}$ promotes keeping the object between the robot and the goal. It is computed as $cos(\alpha) + 1$, where $\alpha$ is the angle between the \textit{robot-object} $(p_R-p_O)$ and \textit{goal-object} $(p_G-p_O)$ vectors and $+1$ is added to make the cost term always non-negative \cite{Williams2017}:

\begin{eqnarray}
    C_{push \ align} = \omega_a\big(\underbrace{\frac{(p_R-p_O)\cdot(p_G - p_O)}{||p_R - p_O||||p_G - p_O||}}_{cos(\alpha)} +1 \big).
\end{eqnarray}
We promote the end-effector to maintain a downward orientation at height $d_h$ using pitch $\theta$ and roll $\phi$:
\begin{equation}
    \nonumber
    C_{ee \ align} = \omega_{ee_r}||[\phi, \theta] - [0, 0]|| + \omega_{ee_h}||p_{R_z} - d_h||.
\end{equation}
The cost is minimized when the end-effector is close to the block at a certain height and orientation, and the block is between the end-effector and the goal at the desired goal pose\footnote{Tuning: $\omega_t = 1$, $\omega_{O_p} = 16$, $\omega_{O_r} = 2$, $\omega_{ee_h} = 8 $, $\omega_{ee_r} = 0.5 $, $\omega_a = 0.8$, $dt = 0.04$, $T = 8$, $K = 500$.}.

We perform the same task as in \cite{arruda_uncertainty_2017} and compare the final results of pushing a squared object on a table surface to two poses (Pose 1 and 2) with a robot arm equipped with a stick. In \Cref{tab:comparisonBaxter}, we report our findings, with our method showing double the accuracy. Our approach performs continuous pushes, unlike the baseline that stops for replanning after each short push. Thus, we complete either task in approximately 8 seconds, while the baseline takes approximately 4 minutes. We used the same evaluation metric of \cite{arruda_uncertainty_2017} for the final cost that is a weighted average of position and orientation errors: $1.5(|p_{G_x} - p_{O_x}| + |p_{G_y} - p_{O_y}|) + 0.01|\psi_O - \psi_G|$. For every run, the object is also randomized in the same way as the rollouts. See the accompanying \href{https://autonomousrobots.nl/paper_websites/isaac-mppi}{video} for the actual behavior. 
\begin{table}[ht]
\medskip
\caption{Summary of comparison with \cite{arruda_uncertainty_2017}} \centering 
\begin{tabular}{C{1.8cm} C{1.5cm} C{1.5cm}} %
\hline\hline 
 \textbf{Approach} & {\textbf{Start pose}} & \textbf{Final cost}\\ [0.5ex] 

\hline 
\multirow{2}{*}{Baseline~\cite{arruda_uncertainty_2017}} & \multicolumn{1}{c}{Pose 1} & 0.057 \\\cline{2-3} 
                                                       & \multicolumn{1}{c}{Pose 2} & 0.079 \\
\hline
\multirow{2}{*}{Ours (sim)} & \multicolumn{1}{c}{Pose 1} & \textbf{0.029 $\pm 0.09$} \\\cline{2-3} 
        & \multicolumn{1}{c}{Pose 2} & \textbf{0.03 $\pm 0.12$} \\ 
\hline
\end{tabular}
\label{tab:comparisonBaxter}
\end{table}




We further compare our approach with \cite{cong_self-adapting_2020} in terms of the success rate of non-prehensile manipulation. Particularly,  we consider the same task settings, pushing 5 different objects to 3 different goal poses. To do so we simply change the objects in the simulation and slightly re-tune the MPPI\footnote{Tuning: $\omega_t = 5$, $\omega_{O_p} = 25$, $\omega_{O_r} = 21$, $\omega_{ee_h} = 30 $, $\omega_{ee_r} = 0.3 $, $\omega_a = 45$, $dt = 0.04$, $T = 8$, $K = 500$.}. 

Since the trained models from \cite{cong_self-adapting_2020} are not provided, we only compare the final results, reported in \Cref{tab:comparisonUR5}. Again, thanks to the continuous pushes, our method takes about 3 seconds per task, while the baseline needs about 24 seconds. We performed 10 pushes per object, totaling 150 pushes. For the non-prehensile manipulation task with the robot arm, the mass and friction of manipulated objects have 30\% uncertainty, and table friction has 90\% uncertainty on the nominal value, sampled uniformly. Size is randomized with zero-mean additive Gaussian noise with a 2 mm standard deviation.
\begin{table}[ht]
\caption{Summary of comparison with \cite{cong_self-adapting_2020}} \centering 
\begin{tabular}{C{1.9cm} C{1.5cm} C{0.5cm} C{0.5cm} C{0.5cm} C{0.5cm} C{0.5cm}} %
\hline\hline 
&  & \multicolumn{5}{c}{\textbf{Object}} \\  \cline{3-7}
\vspace{3mm}\textbf{Approach} & \vspace{3mm} \textbf{Metric} & {{A}} & {B}& {C} & {D} & {E}  \\ [0.5ex] 
\hline 
Baseline \cite{cong_self-adapting_2020} & \multicolumn{1}{c}{Success [\%]} & 93.5 & 90.9 & 93.9 & 91.6 & \textbf{89.5} \\
\hline
Ours (sim)& \multicolumn{1}{c}{Success [\%]}& \textbf{100} & \textbf{93.3} & \textbf{96.7} & \textbf{100} & 66.7 \\
\hline
 \end{tabular}
\label{tab:comparisonUR5}
\end{table}

Our method outperforms both baselines in terms of time to completion, accuracy, and success rate, except for one manipulated object. We achieve this without limiting the sampling to 2D end-effector trajectories, without needing learned models, and without requiring inverse kinematics solvers. 


\subsubsection{Extension to different robots}
\label{subsec:mobile_pushing}
Our method is also easily extensible to different robot platforms and objects because it does not require specialized models or controllers that are robot specific, as opposed to the baselines considered. We chose to use an omnidirectional base, and a differential drive robot, to push a box or a sphere to a goal from different initial configurations. To do so, we only need to change the environment and robot URDF in IsaacGym, and re-tune the cost function for pushing due to different hardware. 

\begin{figure}[htb!]
    \centering
    \medskip
    \includegraphics[width=0.65\linewidth]{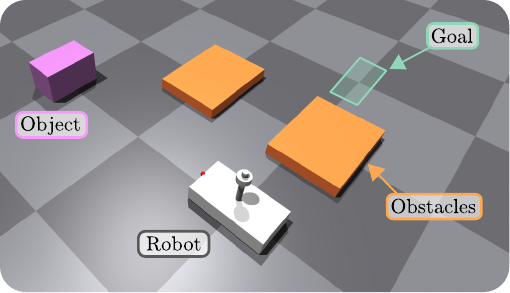}
    \caption{Non-prehensile task using an omnidirectional base. $\omega_t = 0.2$, $\omega_{O_p} = 2$, $\omega_{O_r} = 3$, $\omega_a = 0.6$, $\omega_c = 10$, $T = 8$, $dt = 0.04$, $K = 300$.}
    \label{fig:non_prehensile_heijn_box}
\end{figure}

\paragraph{Omnidirectional push of a box}

The first task is the non-prehensile pushing of a box with an omnidirectional base, see \Cref{fig:non_prehensile_heijn_box}. Success is defined when the box is placed at the goal within 5cm in the $x-y$ direction and within 0.17 radians in rotation. The robot cannot touch obstacles. The cost function for the MPPI is the same as in \Cref{eq:c_push_panda}, re-tuned without considering end effector height and orientation since we now operate on a plane. We add an explicit term for collision avoidance $C_{coll} = \omega_c \sum F_{obst}$:
\begin{equation}
\label{eq:mobile_push}
    C_{push} = C_{dist} + C_{push\ align} + C_{coll},\ 
\end{equation}

\begin{figure}[htb!]
    \centering
    \includegraphics[width=0.65\linewidth]{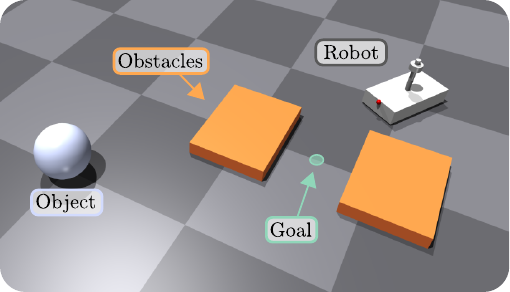}
    \caption{Rolling ball non-prehensile pushing. Goal: Ball placement between two obstacles. $\omega_t = 0.2$, $\omega_{O_p} = 0.1$, $\omega_{O_r} = 0$, $\omega_a = 0.1$, $\omega_c = 0.001$, $T = 8$, $dt = 0.04$, $K = 300$.}
    \label{fig:non_prehensile_heijn_sphere}
\end{figure}

\paragraph{Omnidirectional push of a sphere}
\label{sec:omni_sphere}
One can easily extend the example above to different objects with very different dynamics. We chose a sphere instead of a box, and we simply change the object spawned in the simulation. For this task, we want to put the ball in between the two walls, \Cref{fig:non_prehensile_heijn_sphere}. We considered multiple runs from two different starting poses, A and B. Results are summarized in \Cref{tab:mobile-comp} and the execution can be seen in the accompanying \href{https://autonomousrobots.nl/paper_websites/isaac-mppi}{video}.
\vspace{-0.3cm}

\begin{table}[htb!]
\caption{Results with omnidirectional base}\centering 
\begin{tabular}{C{0.8cm} C{1.0cm} C{0.7cm} C{2.0cm}} %
\hline\hline 
 \textbf{Obj} & \textbf{Env.} & \textbf{Runs} & \textbf{Time [s]} \\ [0.5ex] 
\hline
\multirow{2}{*}{Box} & \multicolumn{1}{c}{Pose A} & 5  & 9.66 $\pm$ 0.84 \\\cline{2-4} 
            & \multicolumn{1}{c}{Pose B} & 5  & 12.84 $\pm$ 0.564\\
\hline
\multirow{2}{*}{Sphere} & \multicolumn{1}{c}{Pose A} & 5  & 8.76 $\pm$ 0.38\\\cline{2-4} 
            & \multicolumn{1}{c}{Pose B} & 5  & 7.45 $\pm$ 0.59 \\
\hline
\end{tabular}
\label{tab:mobile-comp}
\end{table}

\begin{figure}[htb!]
    \centering
    \includegraphics[width=0.65\linewidth]{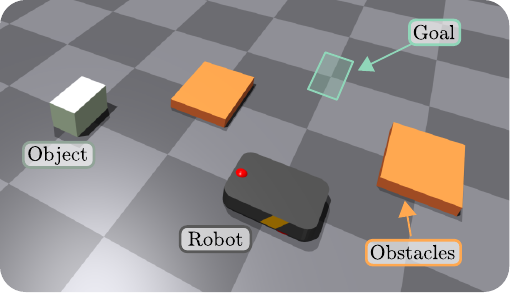}
    \caption{Non-prehensile differential drive pushing. Same task as omnidirectional base. $\omega_t = 0.1$, $\omega_{O_p} = 2$, $\omega_{O_r} = 3$, $\omega_a = 0.6$, $\omega_c = 100$, $T = 12$, $dt = 0.04$, $K = 400$.}
    \label{fig:non_prehensile_boxer}
\end{figure}

\begin{figure*}[!t]
\centering
\medskip
\begin{subfigure}[c]{0.8\textwidth}
   \includegraphics[width=1\linewidth]{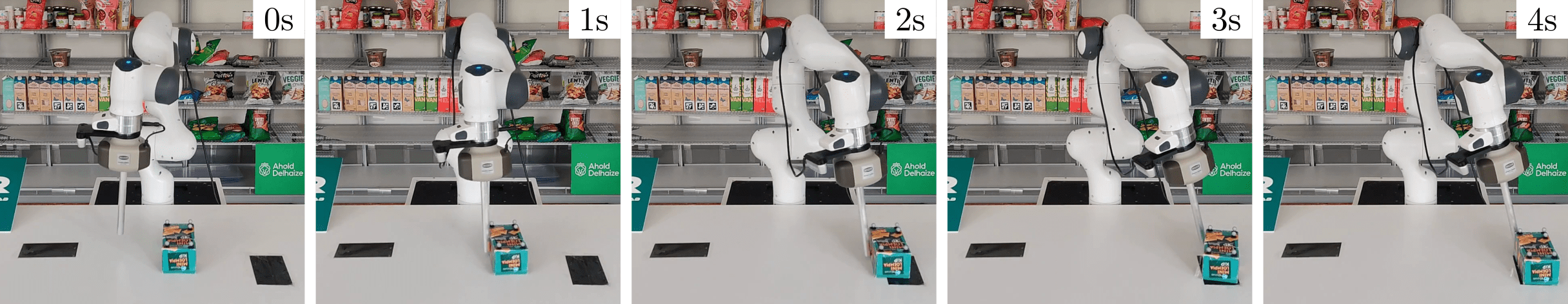}
   \caption{Pushing straight to the goal on the right.}
   \label{fig:realpush_straight} 
\end{subfigure}

\begin{subfigure}[c]{0.8\textwidth}
    \medskip
   \includegraphics[width=1\linewidth]{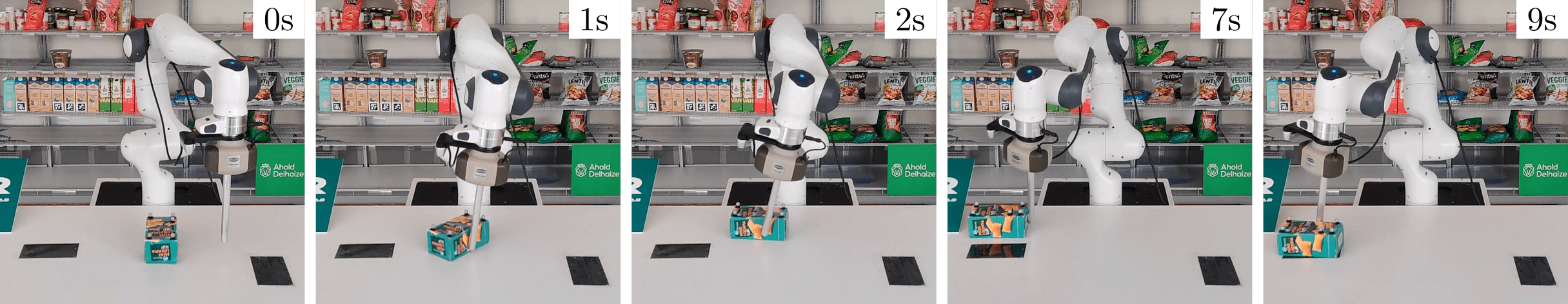}
   \caption{Pushing to the goal on the left with 90$^\circ$ rotation.}
   \label{fig:realpush_90}
\end{subfigure}

\caption{Pushing to two goals with a 7 DOF manipulator directly controlling all joint velocities. Our method allows the end-effector to re-position around the object without specifying any desired contact point.}
   \label{fig:realpush_manipulator} 
\end{figure*}
\paragraph{Differential drive non-prehensile pushing}
We perform differential drive non-prehensile pushing, see \Cref{fig:non_prehensile_boxer}, with the same cost function as before (see \Cref{eq:mobile_push}) but re-tuned. One can change the robot for the task by changing the URDF, neglecting all the additional contact modeling required in a classical model-based MPC. The time taken to push the box to the goal was 18.31s. In the mobile non-prehensile pushing experiments, objects to manipulate are spawned with 30\% uncertainty on mass and friction sampled uniformly, while object size is randomized with Gaussian noise with a standard deviation of 5mm.

\subsection{Real-world experiments}
To demonstrate the applicability of our approach, we transfer to the real world a subset of the non-prehensile manipulation tasks previously presented in~\Cref{subsec:nonprehensile_sim} with both the robot manipulator and the omnidirectional base.
In particular, in \Cref{fig:realpush_manipulator}, we show the results of the 7 DOF manipulator pushing a product to two different goals, similar to the simulations corresponding to \Cref{tab:comparisonBaxter}.
As presented in \Cref{fig:general_idea}, the samples are rolled out in $K=500$ simulated environments in IsaacGym, which, at each timestep, are initialized to the state of the real world. Based on this, the optimal control is estimated and applied to the real system.

When transferring to the real world, compared to the experiments in \Cref{subsec:baselines_pushing}, only the cost function weights were re-tuned. The horizon, control frequency, number of samples, structure of the cost function, and randomization of the sampled environments remained unchanged.

From the experimental evaluation on the real robot, we observe that the time to complete the pushing tasks and the final position errors are comparable to the results in the simulation from \Cref{tab:comparisonBaxter}. Importantly, these results are achieved without making assumptions on specific contact points. Thus, the robot can naturally re-position itself and change contact location autonomously. Additionally, our method allows us to sample joint velocities directly; thus, we do not restrict the sampling to 2D end-effector trajectories to be translated into joint commands, as often seen in other approaches. 

\begin{figure}[htb!]
  \centering
  \begin{subfigure}[c]{0.45\linewidth}
    \centering
    \includegraphics[height=2.8cm]{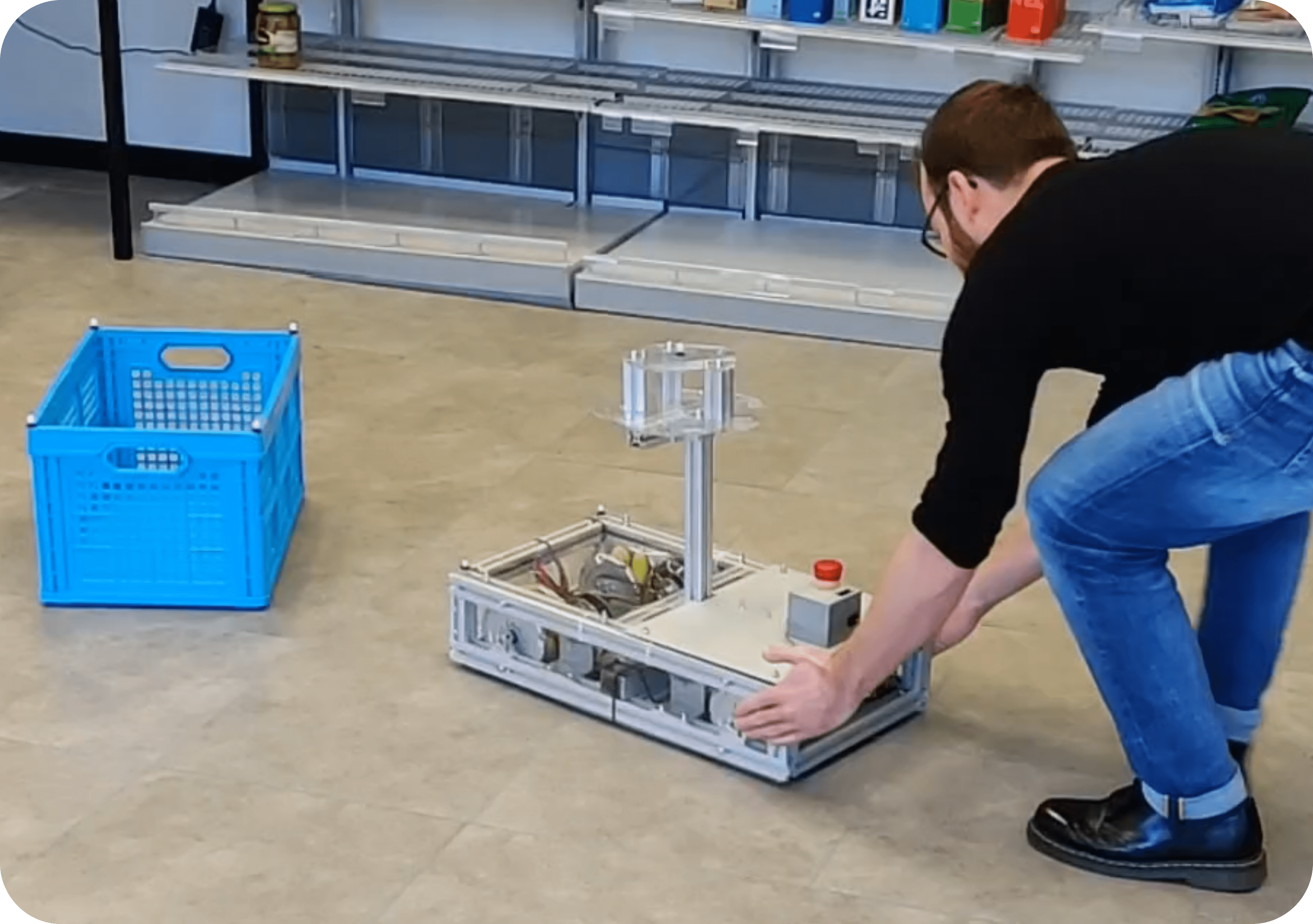}
  \end{subfigure}%
  \begin{subfigure}[c]{0.45\linewidth}
    \centering
    \includegraphics[height=2.8cm]{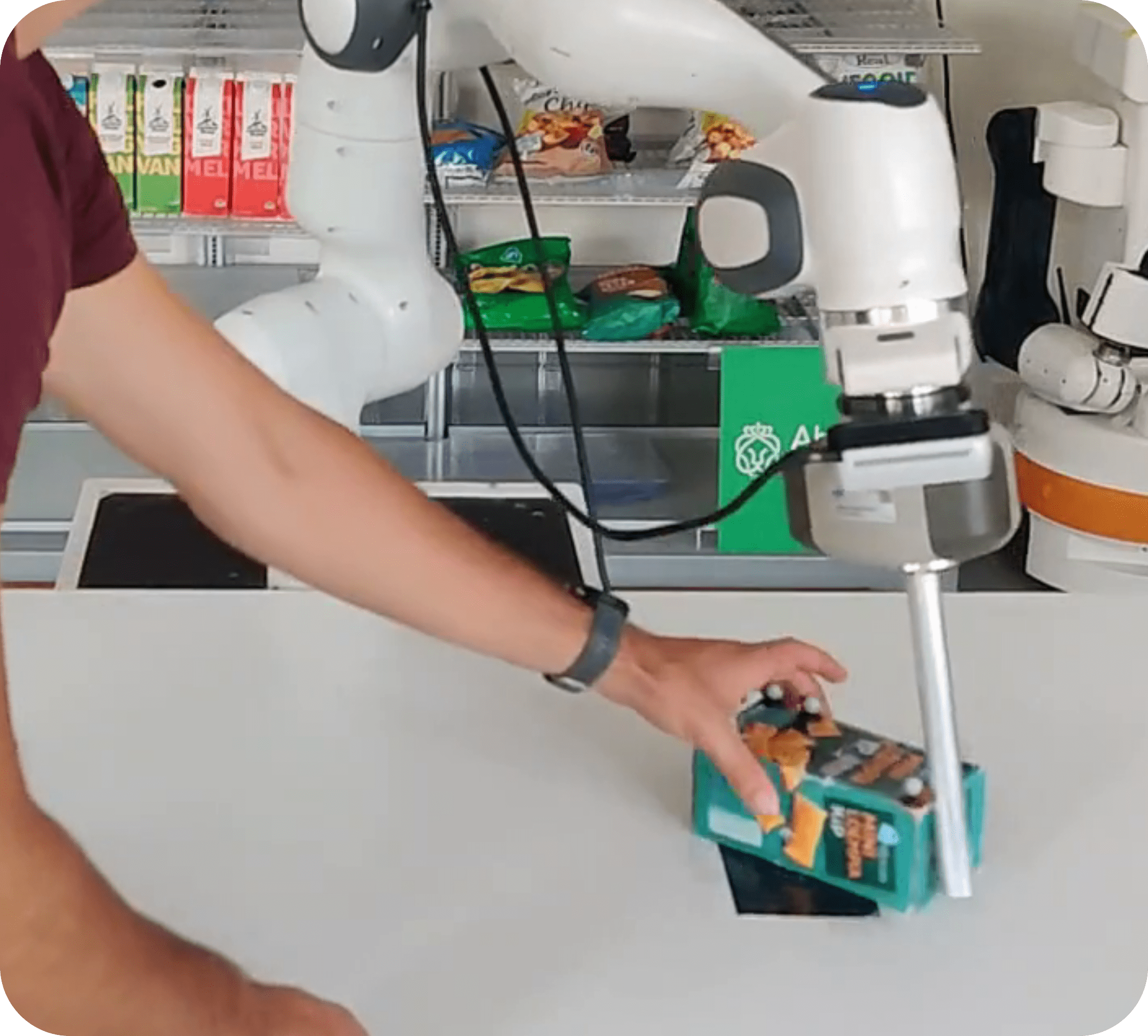}
  \end{subfigure}
    \caption{Qualitative real-world experiments with disturbances. The behavior can be seen in the accompanying \href{https://autonomousrobots.nl/paper_websites/isaac-mppi}{video}.}%
  \label{fig:realworld_disturbed}
\end{figure}

Lastly, to demonstrate robustness, we disturb the execution of pushing tasks by hand with the manipulator and the omnidirectional base (\Cref{fig:realworld_disturbed}). Since we do not assume the robot to be behind the object to be pushed for successful execution, and since the planning and execution happen in real-time at $25Hz$, we can largely perturb the task and let the robot compensate.

\section{Discussion}
\label{sec:discussion}
In this section, we discuss key aspects and potential future work related to our solution. 
First, the computational demands of planning and control with our method can be high when extending the time horizon to several seconds. To keep the time horizon limited for real-time control while preventing being trapped in local minima, future work should incorporate global planning techniques such as A*, RRT, and Probabilistic Roadmaps (PRM) \cite{karaman_sampling-based_2011} to guide the local planner. Similarly to warm starting predictive controllers \cite{mansard_using_2018,lembono_memory_2020}, one could make use of motion libraries of previous executions or learned policies along with random rollouts, to improve the sampling efficiency and exploration~\cite{trevisan_biased_2024}.
Second, in real-world scenarios, uncertainties and discrepancies between simulated and actual environments could present challenges for achieving precise movements and manipulation. We utilized randomization of object properties in the rollouts to address some uncertainties. However, online system identification to converge to the true model parameters is not performed. Enhancing the robustness of the MPPI algorithm itself by reducing model uncertainty, as demonstrated by \cite{abraham_model-based_2020}, could further improve performance. 
Third, tuning control algorithms for optimal performance is time-consuming. Implementing autotuning techniques can automate the process and reduce manual effort. Finally, incorporating additional sensor support, such as lidars and signed distance fields, could be beneficial.

\section{Conclusions}
\label{sec:conclusions}

We presented a way to perform Model Predictive Path Integral controller (MPPI) that uses a physics simulator as the dynamic model. By leveraging the GPU-parallelizable IsaacGym simulator for parallel sampling of forward trajectories, we have eliminated the need for explicit encoding of robot dynamics, contacts, and rigid-body interactions for MPPI. This makes our method easily adaptable to different objects and robots for a wide range of contact-rich motion-planning tasks. Through a series of simulations and real-world experiments, we have demonstrated the effectiveness of this approach in various scenarios, including motion planning with collision avoidance, non-prehensile manipulation, and whole-body control. We showed how our method can compete with state-of-the-art motion planners in case of no interactions, and how it outperforms by a margin other approaches for contact-rich tasks. In addition, we provided an open-source implementation that can be used to reproduce the presented results, and that can be adapted to new tasks and robots. 



\bibliographystyle{IEEEtran}
\bibliography{IEEEabrv,resub}

\end{document}